\documentclass[conference]{IEEEtran}
\IEEEoverridecommandlockouts
\usepackage{cite}
\usepackage{amsmath,amssymb,amsfonts}
\usepackage{algorithmic}
\usepackage{graphicx}
\usepackage{textcomp}
\usepackage{xcolor}
\usepackage{booktabs}
\usepackage{tikz}
\usetikzlibrary{arrows.meta,positioning,fit,calc,shapes.geometric}
\def\BibTeX{{\rm B\kern-.05em{\sc i\kern-.025em b}\kern-.08em
    T\kern-.1667em\lower.7ex\hbox{E}\kern-.125emX}}

\newcommand{\R}{\mathbb{R}}
\newcommand{\dmod}{d}
\newcommand{\dhead}{d_{h}}
\newcommand{\dinner}{d_{\mathrm{ff}}}
\newcommand{\silu}{\mathrm{SiLU}}
\newcommand{\softmax}{\mathrm{softmax}}
\newcommand{\rmsnorm}{\mathrm{RMSNorm}}

\begin{document}

\title{Wiola 13M, a Gated Spiral Attention Architecture for Parameter Efficient Small Language Models}

\author{\IEEEauthorblockN{\textsuperscript{} Aryuemaan Kumar Chowdhury}
\IEEEauthorblockA{\textit{Wiola, OSCOWL ai} \\
\textit{IIT Hyderabad}\\
Hyderabad, India \\
aryu@oscowl.in}
\and
\IEEEauthorblockN{\textsuperscript{} Praveen Oosa}
\IEEEauthorblockA{\textit{Wiola, OSCOWL ai} \\
Hyderabad, India \\
}
\and
\IEEEauthorblockN{\textsuperscript{} Vineesha Reddy}
\IEEEauthorblockA{\textit{Wiola, OSCOWL ai} \\
Hyderabad, India \\
}
}

\maketitle

\begin{abstract}
Small language models in the ten to one hundred million parameter range are attractive for on device inference, rapid experimentation, and controlled scientific study, yet most of them reuse the standard transformer block without adaptation to the small scale regime. We present Wiola, a decoder only language model whose novelty is concentrated in three drop in components of every layer. First, Spiral Rotary Positional Encoding perturbs the standard rotary frequencies by a slowly growing per dimension factor so that phase trajectories fan outward, improving long range discrimination while adding no parameters. Second, Gated Spiral Attention introduces a per head, content adaptive scalar gate derived from a causal cumulative statistic of the query stream, providing an implicit and differentiable form of soft head selection at negligible cost. Third, the Butterfly feed forward block replaces the conventional expansion layer with a multiplicative interaction and an intra block bypass path, matching the parameter count of a four times gated linear unit block while improving gradient flow in shallow stacks. We formalize each component, derive exact parameter and computation budgets, and prove that the gated attention admits an exact and numerically verified equivalence between full sequence training and cached autoregressive decoding, so that no approximation is introduced at inference time. We also describe a fully reproducible training and evaluation protocol on a standard tiny story corpus. The reference implementation is released as an open source package with weights ready publishing support.
\end{abstract}


\section{Introduction}
Large language models have driven rapid progress in natural language processing, but their computational and memory requirements place them out of reach for on device deployment and for laboratories with modest compute budgets. This has renewed interest in the small language model regime, roughly ten to one hundred million parameters, where a model can be trained on a single consumer accelerator in hours and can run locally with a small memory footprint. Work on tiny corpora has shown that even models below ten million parameters can produce fluent and coherent text when the data distribution is controlled \cite{b8}, which makes the small scale regime a productive setting for studying architecture in isolation from data scale.

Despite this interest, the dominant practice is to scale down a standard transformer block unchanged, inheriting design choices that were tuned for models two or three orders of magnitude larger. In the small regime, three properties matter disproportionately. Positional information must be encoded efficiently because sequences are short and every representational dimension is scarce. Attention heads are few, so the failure of even one head to specialize wastes a large fraction of capacity. Finally, the feed forward block dominates the non embedding parameter budget, so its parameter efficiency largely determines the compute optimal shape of the network.

We introduce Wiola, a decoder only model that addresses these three pressures with three targeted modifications to the transformer layer while leaving the surrounding training recipe standard. Our contributions are as follows.
\begin{itemize}
\item \textbf{Spiral Rotary Positional Encoding}, a parameter free modification of rotary position encoding \cite{b2} in which the rotation frequencies are scaled by a slowly increasing factor across dimension pairs, so that phase trajectories fan outward with depth in the frequency spectrum. The modification recovers standard rotary encoding exactly in a limiting case.
\item \textbf{Gated Spiral Attention}, a per head scalar gate computed from a causal cumulative mean of the query projections and applied multiplicatively to the pre softmax scores. The gate lets unhelpful heads suppress themselves, an implicit soft head selection, and it adds only a few hundred parameters per layer.
\item \textbf{The Butterfly feed forward block}, a multiplicative expansion with an intra block bypass path that matches a four times gated linear unit block in parameter count while improving gradient propagation in shallow networks.
\end{itemize}
Beyond the architecture, we provide exact parameter and computation budgets, and we prove and numerically verify that the gated attention preserves an exact equivalence between the parallel training path and the cached decoding path. This last property is important in practice because a naively defined attention gate can silently break the identity between training and generation, degrading sample quality.

\section{Related Work}
\subsection{Positional Encoding}
The original transformer injected order through fixed sinusoidal embeddings \cite{b1}. Rotary Position Encoding \cite{b2} instead rotates queries and keys by an angle proportional to absolute position, so that attention scores depend only on relative offset. Rotary encoding has become the default in modern decoder only models \cite{b5} because it extrapolates gracefully and requires no learned parameters. Alternative schemes such as additive linear biases \cite{b7} also target length extrapolation. Our Spiral encoding stays within the rotary family and modifies only the frequency schedule, preserving the relative offset property while altering how phases separate across the spectrum.

\subsection{Feed Forward and Gated Linear Units}
Gated linear units and their variants improve the transformer feed forward block by introducing a multiplicative interaction between two projections \cite{b3,b12}. These variants typically use three projection matrices and reduce the inner width to hold parameters constant. The Butterfly block follows this tradition but adds an explicit intra block bypass, which we find useful for gradient flow when the number of layers is small.

\subsection{Attention Efficiency and Gating}
A large body of work reduces attention cost by sharing keys and values across heads \cite{b6} or by restructuring the attention computation. These methods target the large model regime and the memory bandwidth of long contexts. Our goal is different. Rather than reduce the cost of attention, we make each of a small number of heads more useful by allowing a content adaptive gate to modulate its contribution, which is closer in spirit to conditional computation than to attention compression.

\subsection{Small Language Models}
The tiny story setting \cite{b8} demonstrated that small models can learn to generate coherent text from a controlled distribution, and compute optimal scaling analysis \cite{b9} clarified how parameters and tokens should be balanced. Wiola targets exactly this regime and is designed to be a clean and reproducible baseline within it.

\section{Model Architecture}
\subsection{Overview and Notation}
\begin{figure}[t]
    \centering
    \includegraphics[width=0.42\textwidth]{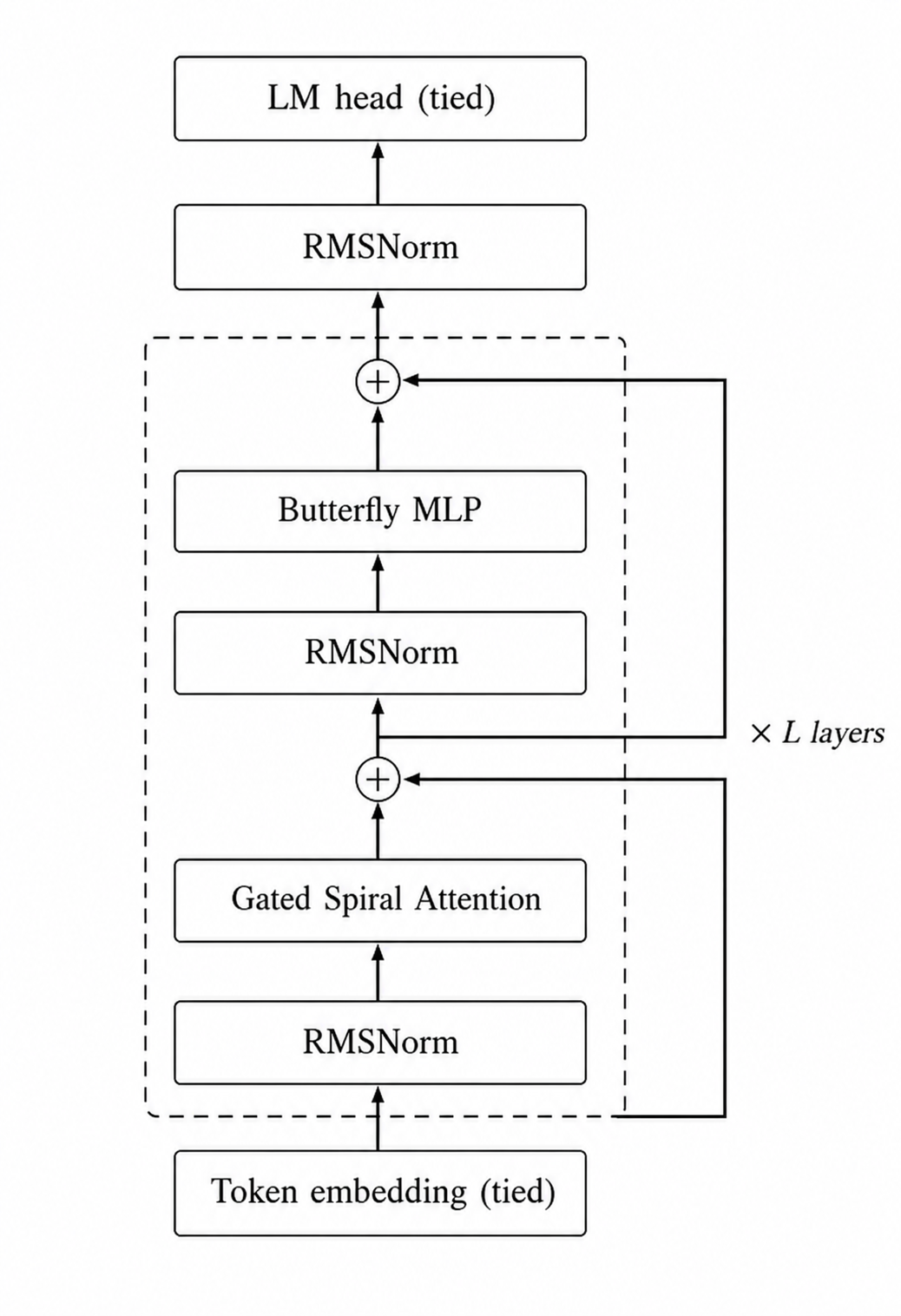} 
    \caption{Wiola decoder. Each of the $L$ layers applies a pre-norm Gated Spiral Attention sub-block and a pre-norm Butterfly feed-forward sub-block, each wrapped in a residual connection. The input embedding and the output projection share weights.}
    \label{fig:overview}
\end{figure}

\begin{figure*}[t]
    \centering
    \includegraphics[width=0.92\textwidth]{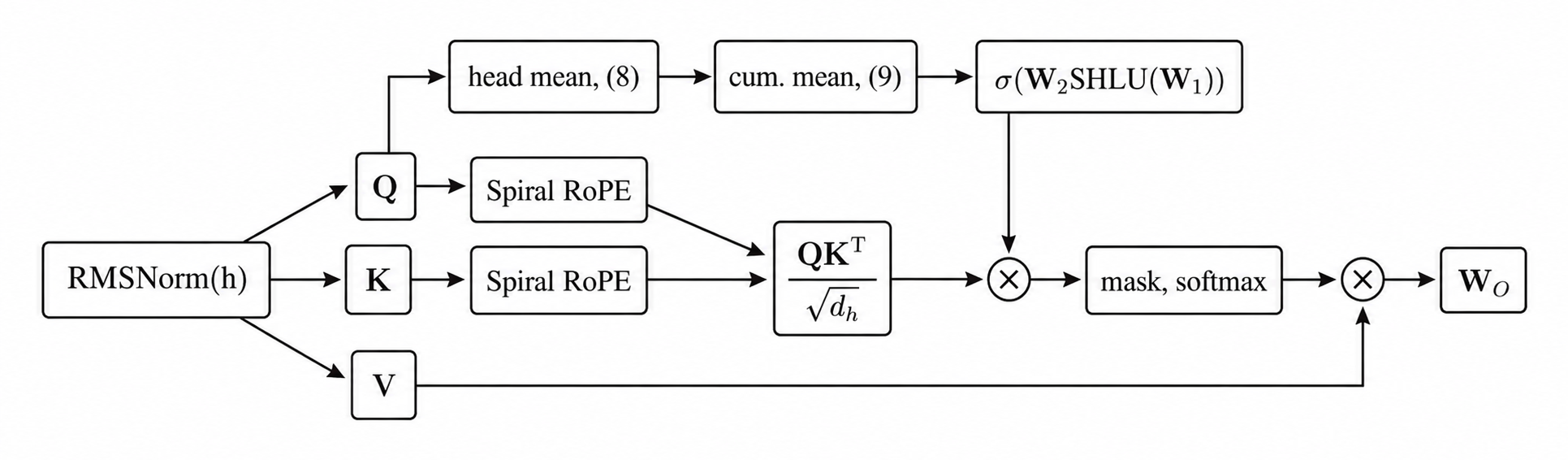} 
    \caption{Gated Spiral Attention. Queries and keys receive Spiral encoding. A parallel branch derives a per-head gate from a causal cumulative mean of the queries and multiplies it into the scores before the softmax. The running cumulative sum is the only additional quantity that must be cached during decoding.}
    \label{fig:gsa}
\end{figure*}

Wiola is a pre norm decoder only transformer. Let $\dmod$ denote the hidden width, $L$ the number of layers, $H$ the number of attention heads, and $\dhead=\dmod/H$ the per head width. A token sequence of length $T$ is embedded, processed by $L$ identical decoder layers, normalized, and projected to vocabulary logits by a head whose weights are tied to the input embedding. Each layer applies two residual sub blocks,
\begin{align}
\mathbf{h}' &= \mathbf{h} + \mathrm{GSA}\!\left(\rmsnorm(\mathbf{h})\right), \label{eq:res1}\\
\mathbf{h}'' &= \mathbf{h}' + \mathrm{FFN}\!\left(\rmsnorm(\mathbf{h}')\right), \label{eq:res2}
\end{align}
where $\mathrm{GSA}$ is Gated Spiral Attention and $\mathrm{FFN}$ is the Butterfly block. Root mean square normalization \cite{b4} is used throughout,
\begin{equation}
\rmsnorm(\mathbf{x}) = \frac{\mathbf{x}}{\sqrt{\tfrac{1}{\dmod}\sum_{j=1}^{\dmod} x_j^2 + \varepsilon}}\odot\boldsymbol{\gamma}, \label{eq:rms}
\end{equation}
with a learned gain $\boldsymbol{\gamma}\in\R^{\dmod}$ and a small constant $\varepsilon$. Fig.~\ref{fig:overview} shows the overall data flow, and Fig.~\ref{fig:gsa} details the attention block.

\subsection{Spiral Rotary Positional Encoding}
\begin{figure}[t]
\centering
\includegraphics[width=0.98\linewidth]{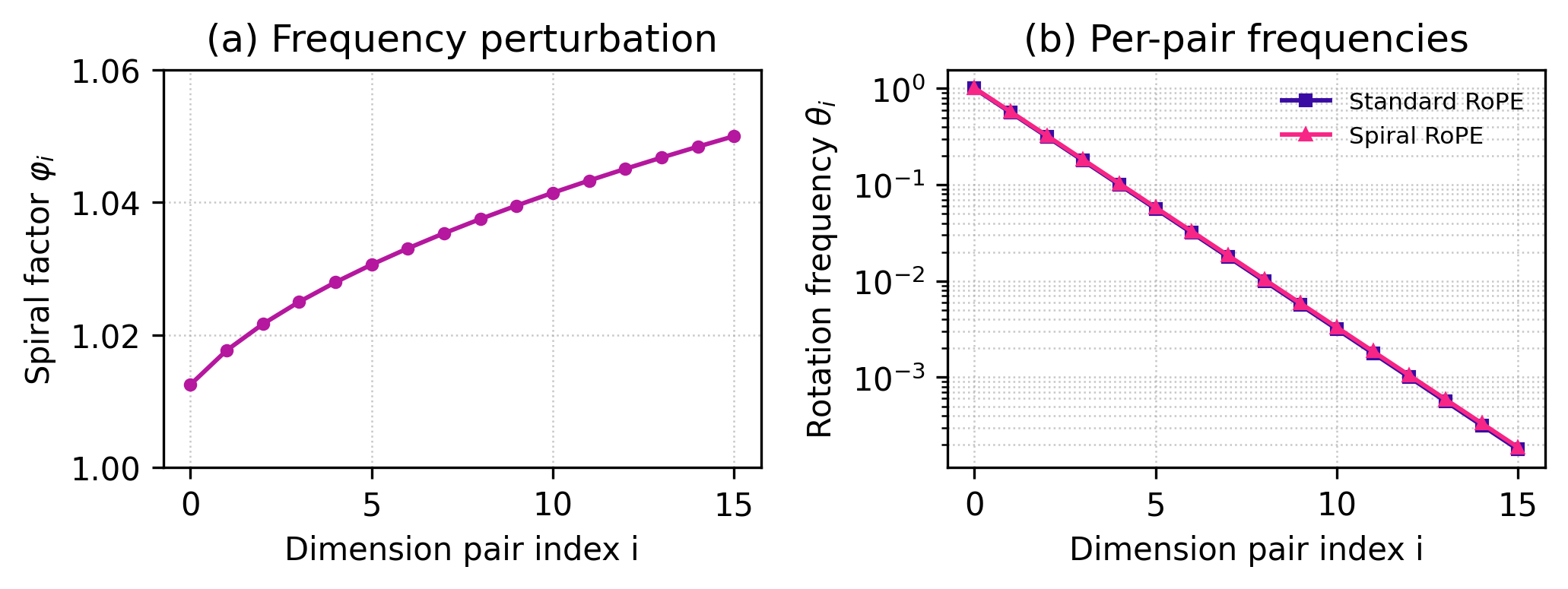}
\caption{Spiral Rotary Positional Encoding for head width $\dhead=32$ and $\alpha=0.05$. Panel (a) shows the parameter free multiplicative factor $\varphi_i$ in \eqref{eq:phi}, which grows monotonically from about $1.0125$ to $1.05$. Panel (b) compares standard and spiral frequencies on a logarithmic axis. The perturbation is largest for the higher frequency pairs.}
\label{fig:spiral}
\end{figure}

Standard rotary encoding \cite{b2} partitions each head vector into $\dhead/2$ coordinate pairs and rotates pair $i$ at position $m$ by an angle $m\theta_i$, with a geometric frequency schedule
\begin{equation}
\theta_i = \mathrm{base}^{-2i/\dhead}, \qquad i=0,\dots,\tfrac{\dhead}{2}-1, \label{eq:theta}
\end{equation}
where $\mathrm{base}=10^{4}$. Spiral encoding scales each frequency by a slowly growing, parameter free factor
\begin{align}
\varphi_i &= 1 + \alpha\,\frac{\sqrt{i+1}}{\sqrt{\dhead/2}}, \label{eq:phi}\\
\tilde{\theta}_i &= \theta_i\,\varphi_i, \label{eq:thetatilde}
\end{align}
with a single fixed hyperparameter $\alpha$. The factor grows monotonically from $1$ at the lowest pair toward $1+\alpha$ at the highest pair, so higher frequency pairs receive a proportionally larger boost and their phases separate more rapidly with position, while the lowest pairs remain almost unchanged and preserve local continuity. Setting $\alpha=0$ recovers \eqref{eq:theta} exactly, so Spiral encoding is a strict generalization of rotary encoding. Fig.~\ref{fig:spiral} plots the factor and the resulting frequencies for the head width used by our smallest configuration.

With $\Theta_{m,i}=m\tilde{\theta}_i$, the rotation applied to the coordinate pair $(2i,2i+1)$ of a query or key vector $\mathbf{u}$ is
\begin{equation}
\begin{bmatrix} u'_{2i}\\ u'_{2i+1}\end{bmatrix}
=
\begin{bmatrix} \cos\Theta_{m,i} & -\sin\Theta_{m,i}\\[2pt] \sin\Theta_{m,i} & \cos\Theta_{m,i}\end{bmatrix}
\begin{bmatrix} u_{2i}\\ u_{2i+1}\end{bmatrix}. \label{eq:rot}
\end{equation}
Because the transformation is a per position rotation applied identically to queries and keys, the dot product between a rotated query at position $t$ and a rotated key at position $s$ remains a function of the relative offset $t-s$, so the relative offset property of rotary encoding is preserved.

\subsection{Gated Spiral Attention}
Let $\mathbf{X}\in\R^{T\times\dmod}$ be the normalized layer input. Queries, keys, and values are formed by linear projections and split into $H$ heads, $\mathbf{Q}^{(h)},\mathbf{K}^{(h)},\mathbf{V}^{(h)}\in\R^{T\times\dhead}$. Spiral encoding \eqref{eq:rot} is applied to the queries and keys.

\paragraph{Causal gate}
The gate summarizes the query stream up to the current position through a head averaged context vector and its causal cumulative mean,
\begin{align}
\mathbf{m}_s &= \frac{1}{H}\sum_{h=1}^{H}\mathbf{q}^{(h)}_s \in\R^{\dhead}, \label{eq:mavg}\\
\mathbf{c}_t &= \frac{1}{t}\sum_{s=1}^{t}\mathbf{m}_s \in\R^{\dhead}. \label{eq:cmean}
\end{align}
A two layer gating network then produces one scalar per head,
\begin{equation}
\mathbf{g}_t = \sigma\!\big(\mathbf{W}_2\,\silu(\mathbf{W}_1\mathbf{c}_t)+\mathbf{b}_2\big)\in(0,1)^{H}, \label{eq:gate}
\end{equation}
with $\mathbf{W}_1\in\R^{H\times\dhead}$ and $\mathbf{W}_2\in\R^{H\times H}$. The bias $\mathbf{b}_2$ is initialized to zero, so at the start of training every gate is approximately one half and attention is close to the ungated baseline. Because $\mathbf{c}_t$ depends only on positions $s\le t$, the gate is strictly causal.

\paragraph{Gated scores}
For head $h$, the gate multiplies the scaled scores before the softmax,
\begin{align}
A^{(h)}_{t,s} &= g^{(h)}_t\,\frac{\big\langle \mathbf{q}^{(h)}_t,\,\mathbf{k}^{(h)}_s\big\rangle}{\sqrt{\dhead}} + M_{t,s}, \label{eq:scores}\\
\mathbf{z}^{(h)}_t &= \sum_{s=1}^{T}\softmax_s\!\big(A^{(h)}_{t,\cdot}\big)\,\mathbf{v}^{(h)}_s, \label{eq:av}
\end{align}
where $M$ is the causal mask with $M_{t,s}=0$ for $s\le t$ and $-\infty$ otherwise. The per head outputs are concatenated and projected,
\begin{equation}
\mathrm{GSA}(\mathbf{X})_t = \big[\mathbf{z}^{(1)}_t\,\Vert\cdots\Vert\,\mathbf{z}^{(H)}_t\big]\,\mathbf{W}_O. \label{eq:o}
\end{equation}
Applying the gate before the softmax means that when $g^{(h)}_t$ is small the row is flattened toward a near uniform distribution and the head contributes little, and gradient pressure drives the gates of consistently unhelpful heads toward zero. This is an implicit and fully differentiable soft head selection that requires no auxiliary sparsity loss. By default the gate is computed from the pre rotation queries so that it depends on content rather than absolute position, and this choice is exposed as a configuration flag.

\subsection{Butterfly Feed Forward Block}
The feed forward block expands the hidden state into two halves through a single projection, applies a multiplicative gate, and adds an intra block bypass before the down projection,
\begin{align}
[\mathbf{a}\,\Vert\,\mathbf{b}] &= \mathbf{x}\,\mathbf{W}_{\mathrm{up}}^{\top}, \qquad \mathbf{a},\mathbf{b}\in\R^{\dinner}, \label{eq:up}\\
\mathbf{r} &= \silu(\mathbf{a})\odot\mathbf{b} + \mathbf{W}_{\mathrm{bp}}\,\mathbf{x}, \label{eq:gateff}\\
\mathrm{FFN}(\mathbf{x}) &= \mathbf{W}_{\mathrm{down}}\,\mathbf{r}, \label{eq:down}
\end{align}
where $\mathbf{W}_{\mathrm{up}}\in\R^{2\dinner\times\dmod}$, $\mathbf{W}_{\mathrm{bp}}\in\R^{\dinner\times\dmod}$, and $\mathbf{W}_{\mathrm{down}}\in\R^{\dmod\times\dinner}$. The multiplicative term provides the same expressiveness as gated linear unit feed forward blocks \cite{b3}, while the additive bypass supplies a linear path from input to the block output that improves gradient flow when $L$ is small. With the choice $\dinner=2\dmod$, the three matrices contain $4\dmod\,\dinner$ parameters, identical to a conventional four times expansion block with a single activation, so the added multiplicative and bypass structure comes at no parameter cost relative to that baseline.

\subsection{Training Objective}
The model is trained with the standard next token objective. With tied embedding matrix $\mathbf{E}\in\R^{V\times\dmod}$, the logits are $\boldsymbol{\ell}_t = \rmsnorm(\mathbf{h}^{(L)}_t)\,\mathbf{E}^{\top}$ and the loss is the mean token negative log likelihood,
\begin{equation}
\mathcal{L} = -\frac{1}{T}\sum_{t=1}^{T}\log\frac{\exp\!\big(\ell_{t,x_t}\big)}{\sum_{v=1}^{V}\exp\!\big(\ell_{t,v}\big)}. \label{eq:loss}
\end{equation}

\section{Theoretical Analysis}
\subsection{Parameter Budget}
Table~\ref{tab:variants} lists three reference configurations. The Nano configuration is intended for the tiny story regime, and Micro and Small trade compute for capacity. Table~\ref{tab:params} decomposes the Nano budget. The attention block contributes the four projection matrices, the gate contributes only $H\dhead + H^2 + H$ parameters, and the Butterfly block contributes $4\dmod\dinner$. Summed over six layers and added to the tied embedding, the total is $12{,}915{,}888$ parameters, confirming the design target of about $12.9$ million. The gate accounts for $328$ parameters per layer, four orders of magnitude smaller than either the attention or the feed forward block, which quantifies the claim that the gating mechanism is essentially free.

\begin{table}[t]
\caption{Reference Configurations of Wiola}
\label{tab:variants}
\centering
\footnotesize
\begin{tabular}{lcccccc}
\toprule
\textbf{Variant} & $\dmod$ & $L$ & $H$ & $\dinner$ & \textbf{ctx} & \textbf{Params} \\
\midrule
Nano  & 256 & 6  & 8  & 512  & 1024 & 12.9\,M \\
Micro & 384 & 8  & 12 & 768  & 1024 & $\approx$40\,M \\
Small & 512 & 12 & 16 & 1024 & 2048 & $\approx$90\,M \\
\bottomrule
\end{tabular}
\end{table}

\begin{table}[t]
\caption{Parameter Budget of the Nano Configuration}
\label{tab:params}
\centering
\footnotesize
\begin{tabular}{lcr}
\toprule
\textbf{Component} & \textbf{Expression} & \textbf{Count} \\
\midrule
Attention projections (per layer) & $4\dmod^{2}$ & 262{,}144 \\
Attention gate (per layer)        & $H\dhead+H^{2}+H$ & 328 \\
Butterfly block (per layer)       & $4\dmod\dinner$ & 524{,}288 \\
Normalization (per layer)         & $2\dmod$ & 512 \\
\midrule
Per layer subtotal                & --- & 787{,}272 \\
All six layers                    & $6\times$ & 4{,}723{,}632 \\
Tied token embedding              & $V\dmod$ & 8{,}192{,}000 \\
Final normalization               & $\dmod$ & 256 \\
\midrule
\textbf{Total}                    & --- & \textbf{12{,}915{,}888} \\
\bottomrule
\end{tabular}
\end{table}

\subsection{Computational Complexity}
Table~\ref{tab:complexity} summarizes the per layer cost as a function of sequence length $T$. The projections and the feed forward block are quadratic in width and linear in length, while the score and value products are quadratic in length as in any dense attention. The gate adds only the cost of a cumulative mean and a small two layer network, which is linear in $T$ and independent of $T^2$. The gate therefore does not change the asymptotic cost of the layer, which remains $\mathcal{O}(T\dmod^{2}+T^{2}\dmod)$.

\begin{table}[t]
\caption{Per Layer Computational Cost}
\label{tab:complexity}
\centering
\footnotesize
\begin{tabular}{lc}
\toprule
\textbf{Sub computation} & \textbf{Cost} \\
\midrule
Query key value and output projections & $\mathcal{O}(T\dmod^{2})$ \\
Score and value products               & $\mathcal{O}(T^{2}\dmod)$ \\
Attention gate                         & $\mathcal{O}\!\big(T(\dhead H + H^{2})\big)$ \\
Butterfly feed forward                 & $\mathcal{O}(T\dmod\dinner)$ \\
\bottomrule
\end{tabular}
\end{table}

\subsection{Exact Equivalence of Training and Cached Decoding}
A gate defined through a running statistic can easily break the identity between the parallel training computation and the step by step decoding computation, which would make generated samples diverge from what the training objective rewards. Wiola is designed so that this identity holds exactly.

\vspace{2pt}
\noindent\textbf{Proposition 1.} \textit{For any input sequence, the per head gate $\mathbf{g}_t$ produced by a single full sequence forward pass equals the gate produced by autoregressive decoding that caches the running sum $\mathbf{S}_{t}=\sum_{s\le t}\mathbf{m}_s$. Consequently the two computations yield identical attention scores and identical outputs at every position.}
\vspace{2pt}

\begin{IEEEproof}
By \eqref{eq:mavg} and \eqref{eq:cmean}, $\mathbf{g}_t$ depends on position only through $\mathbf{c}_t=\mathbf{S}_t/t$. In the parallel path, $\mathbf{S}_t$ is the prefix sum of the head averaged queries and $t$ is the position index, both computed directly. In the cached path, decoding maintains $\mathbf{S}_{t}=\mathbf{S}_{t-1}+\mathbf{m}_t$ and the integer count $t$, so it recovers the same $\mathbf{S}_t$ and the same $t$, hence the same $\mathbf{c}_t$ and the same $\mathbf{g}_t$. Key and value caching is exact by construction. Because \eqref{eq:scores} through \eqref{eq:o} are deterministic functions of $\mathbf{g}_t$, the queries, the keys, and the values, identical inputs yield identical outputs.
\end{IEEEproof}

We verified the proposition numerically by comparing a full sequence forward pass against step by step cached decoding on random inputs, with and without the gate active. The two paths agreed to floating point tolerance at every position, confirming that caching introduces no approximation. The only additional state that decoding must carry is the running sum $\mathbf{S}_t$, a vector of $\dhead$ values per layer as shown in Algorithm~\ref{alg:decode}.

\begin{figure}[t]
\centering
\begin{minipage}{0.94\linewidth}
\begin{algorithmic}[1]
\STATE \textbf{state} per layer $\ell$: key cache, value cache, running sum $\mathbf{S}^{\ell}\gets\mathbf{0}$, count $n\gets 0$
\STATE $n\gets n+1$
\FOR{each layer $\ell$}
  \STATE compute $\mathbf{q},\mathbf{k},\mathbf{v}$ for the new token
  \STATE $\mathbf{m}\gets$ head mean of $\mathbf{q}$ \COMMENT{Eq.~\eqref{eq:mavg}}
  \STATE $\mathbf{S}^{\ell}\gets\mathbf{S}^{\ell}+\mathbf{m}$; \quad $\mathbf{c}\gets\mathbf{S}^{\ell}/n$
  \STATE $\mathbf{g}\gets\sigma(\mathbf{W}_2\silu(\mathbf{W}_1\mathbf{c})+\mathbf{b}_2)$
  \STATE apply Spiral encoding to $\mathbf{q},\mathbf{k}$; append $\mathbf{k},\mathbf{v}$ to caches
  \STATE scores $\gets\mathbf{g}\odot(\mathbf{q}\,\mathbf{K}^{\top}/\sqrt{\dhead})$; attend over cached values
\ENDFOR
\RETURN next token logits
\end{algorithmic}
\end{minipage}
\caption{One decoding step with cached gate state. The running sum $\mathbf{S}^{\ell}$ and the count $n$ make the gate identical to the parallel computation.}
\label{alg:decode}
\end{figure}

\subsection{Inference Memory}
During decoding the dominant memory term is the key and value cache, which for a context of length $T$ stores $2LT\dmod$ elements. For the Nano configuration at a context of $1024$ tokens in half precision this is about $6.3$ megabytes. The gate adds only $L\dhead$ elements of running state, which for Nano is $192$ values in total and is therefore negligible relative to the cache.

\section{Reproducible Evaluation Protocol}
Because the contribution of this work is architectural, we specify a fully reproducible protocol so that the analytical results above can be complemented by empirical language modeling numbers under identical conditions. We recommend the tiny story corpus \cite{b8} for the Nano configuration. A byte level tokenizer with a vocabulary of $32{,}000$ is either trained on the corpus or reused from an existing model. Optimization uses AdamW with momentum coefficients $0.9$ and $0.95$, a cosine schedule with a peak learning rate of $3\times10^{-4}$, a sequence length of $512$, and mixed precision. Table~\ref{tab:eval} gives the evaluation template. The two quantities marked as analytical are reported directly from the model definition and do not require training, while the language modeling entries are to be filled from a training run of the released code. We deliberately leave the language modeling cells unspecified rather than report untrained numbers, so that the table reflects only measurements that have actually been obtained.

\begin{table}[t]
\caption{Evaluation Template for the Nano Configuration}
\label{tab:eval}
\centering
\footnotesize
\begin{tabular}{lc}
\toprule
\textbf{Metric} & \textbf{Value} \\
\midrule
Parameters (analytical) & 12{,}915{,}888 \\
Key value cache at 1024 tokens, fp16 (analytical) & $\approx 6.3$\,MB \\
Validation perplexity & to be reported \\
Tiny story graded score & to be reported \\
Tokens per second, single accelerator & to be reported \\
\bottomrule
\end{tabular}
\end{table}

An ablation plan follows directly from the three components. Setting $\alpha=0$ isolates the effect of Spiral encoding against standard rotary encoding. Disabling the gate isolates the effect of Gated Spiral Attention. Replacing the Butterfly block with a conventional gated linear unit block of equal parameter count isolates the effect of the bypass path. Because all three toggles are exposed in the released configuration, each ablation is a single flag change.

\section{Discussion and Limitations}
The analysis establishes that the three components are inexpensive and that the gated attention is exact under caching, but it does not by itself establish an accuracy improvement, which must be measured empirically under the protocol above. Three limitations are worth stating plainly. First, the spiral coefficient $\alpha$ is a fixed hyperparameter rather than a learned quantity, and a learned or per layer schedule may be preferable. Second, the gate uses a first moment of the query stream, and richer causal statistics could capture more context at a still modest cost. Third, the reference implementation uses an explicit attention computation so that the gate can act on the pre softmax scores, which forgoes fused attention kernels and their memory benefits at long context. None of these limitations affects the correctness results, and each suggests a concrete direction for future work.

\section{Conclusion}
We presented Wiola, a small language model whose novelty is concentrated in three drop in layer components. Spiral Rotary Positional Encoding fans out rotary phases at no parameter cost, Gated Spiral Attention performs implicit soft head selection through a causal and cache exact gate, and the Butterfly feed forward block adds multiplicative and bypass structure without increasing the parameter budget. We derived exact parameter and computation budgets, proved and numerically verified the equivalence between training and cached decoding, and specified a reproducible protocol for empirical evaluation. The architecture is released as an open source package to serve as a clean baseline for research in the small language model regime.

\section*{Acknowledgment}
The authors thank the open source community whose libraries for tensor computation and model hosting made the reference implementation possible.

\end{document}